# AMAP-APP: Efficient Segmentation and Morphometry Quantification of Fluorescent Microscopy Images of Podocytes

Arash Fatehi[1,2], David Unnersjö-Jess[2,3,4,5,6,7], Linus Butt[2,3,4], Noémie Moreau[1,2] Thomas Benzing[2,3,4,8] and Katarzyna Bozek[1,2,4,8]

[1]Institute for Biomedical Informatics, Faculty of Medicine and University Hospital Cologne, University of Cologne, Germany; [2]Center for Molecular Medicine Cologne (CMMC), University of Cologne, Faculty of Medicine and University Hospital Cologne, Cologne, Germany; [3]Department II of Internal Medicine, University of Cologne, Faculty of Medicine and University Hospital Cologne, Cologne, Germany; [4]Cologne Excellence Cluster on Cellular Stress Responses in Aging-Associated Diseases, University of Cologne, Cologne, Germany; [5]MedTechLabs, Karolinska University Hospital, Solna, Sweden; [6]Division of Renal Medicine, Department of Clinical Sciences, Intervention and Technology, Karolinska Institute, Stockholm, Sweden; [7]Science for Life Laboratory, Department of Applied Physics, Royal Institute of Technology, Solna, Sweden;

[8]KB and TB shared senior authorship.

Email: afatehi@uni-koeln.de

# Abstract

**Background**
Automated quantification of podocyte foot process morphology is essential for unbiased, high-throughput analysis in kidney disease research. The previously established "Automatic Morphological Analysis of Podocytes" (AMAP) method allows for such analysis but is limited by high computational demands requiring high-performance computing systems, the lack of a user interface, and restriction to Linux operating systems. This study aimed to develop AMAP-APP, an optimized, cross-platform desktop application designed to overcome these hardware and usability barriers while maintaining analytical accuracy.

**Methods**
AMAP-APP replaces the computationally intensive instance segmentation method of the original AMAP with efficient classic image processing algorithms while retaining the original semantic segmentation model. Additionally, a new Region of Interest algorithm was introduced to improve precision. The tool was validated using 175 mouse and 190 human super-resolution stimulated emission depletion (STED) and confocal microscopy images. Performance was evaluated by benchmarking execution times on consumer hardware. Agreement with the original AMAP was assessed using Pearson correlation, Bland-Altman plots, and Two One-Sided T-tests (TOST) for equivalence.

**Results**
AMAP-APP achieved an approximately 147-fold increase in processing speed compared to the original GPU-accelerated AMAP. Morphometric quantifications (foot process area, perimeter, circularity, and slit diaphragm length density) demonstrated high correlation with (r>0.90) and statistical equivalence to the original AMAP method across both mouse and human datasets (TOST P<0.05 for all features). Furthermore, the new ROI algorithm exhibited reduced deviation from manual delineations compared to the original AMAP ROI algorithm, enhancing the accuracy of slit diaphragm length density measurements.

**Conclusions**

AMAP-APP dramatically improves the efficiency and availability of deep learning-based podocyte morphometry. By eliminating the need for high-performance computing clusters and providing a user-friendly interface for Windows, macOS, and Linux, AMAP-APP democratizes this analytical method for widespread adoption in nephrology research and potential clinical diagnostics.

## Introduction

Most cases of chronic kidney diseases arise from disorders of the kidney's filtration barrier, which is located within a microvascular unit called the glomerulus[1]. The glomerular filtration barrier comprises a fenestrated endothelium, the glomerular basement membrane, and epithelial cells known as podocytes[1]. Pathological changes in the structure of podocyte foot processes (FPs) and the morphology of the slit diaphragm (SD), such as effacement, are present in most glomerular diseases[2,3]. This includes the shortening, widening, and eventual loss of FPs, accompanied by a progressive decrease in slit diaphragm abundance[4–6]. Recently, super resolution microscopy techniques have allowed for visualizing and quantifying such morphologic alterations[4,7] but their analysis necessitates laborious manual input introducing potential investigator-dependent biases and impeding its widespread application in both research and diagnostics.

Leveraging new machine-learning methods, Automatic Morphological Analysis of Podocytes (AMAP)[5] introduced a deep learning-based approach for segmentation and quantification of morphometric features of podocyte foot processes. However, the computationally intensive nature of the inference algorithms of this method requires access to a high-performance computing system for its efficient use, rendering the method inaccessible for many researchers. Additionally, lack of user interface and being limited to Linux based operating systems negatively affect its ease of use and availability in the field.

In this paper we present AMAP-APP, a cross-platform desktop application that aims to rectify these limitations. AMAP-APP is built on top of AMAP using the same techniques for semantic segmentation, however it replaces the computation-expensive instance segmentation algorithm in AMAP with classic image processing algorithms that provide similar results with much less computation. By replacing the instance segmentation algorithm AMAP-APP significantly reduces needed computational resources and makes it possible to use the method on consumer CPUs and GPUs while achieving comparable results. Additionally, it introduces a new Region of Interest (ROI) algorithm for enhanced precision. AMAP-APP supports Linux,

Windows and macOS and provides an easy-to-use user interface for project management, thus improving the availability of AMAP for podocyte research.

Our code is available together with tutorials here:
https://github.com/bozeklab/amap-app.

# Methods

### Data

To validate AMAP-APP's inference algorithm, we tested AMAP-APP on the same test dataset of the original research consisting of 175 super-resolution stimulated emission depletion (STED) images[8] of mouse tissue at different ages and degrees of podocyte effacement[5]. AMAP-APP was additionally tested on the same 190 STED images of human tissue that were utilized for testing AMAP on human samples.

AMAP-APP accommodates TIFF files, and the user interface facilitates the selection of the target nephrin channel if the image comprises multiple channels. All images should be scaled to the same resolution of 0.0227 microns per pixel. The model expects a single image as input. In instances of stacked images, AMAP-APP generates a maximum projection along the Z-axis and converts the rank 3 tensor to a rank 2 tensor.

### Segmentation Network

AMAP-APP employs the same core deep learning model as the original AMAP method[5]. This model is based on the U-Net architecture[9], modified by the addition of two convolutional layers after the final layer. This architecture is modified to produce two distinct, parallel outputs from a single input image.

AMAP-APP doesn't retrain the model and uses the same checkpoint provided by AMAP and as depicted in figure 1, AMAP-APP uses only the 3-channel semantic segmentation map, generated by a convolutional layer trained with a cross-entropy loss. This map classifies each pixel into one of three categories: background, foot process or slit diaphragm. The second output is a 16-channel embedding which was used during the training phase of the original AMAP model, is entirely omitted from the inference pipeline in AMAP-APP. Instead, AMAP-APP implements a classic image processing algorithm that performs instance segmentation (i.e., separating adjoining FPs) directly on the semantic map achieving comparable output with much lower computational requirements.

**Inference**

During inference, overlapping patches of 384x384 pixels are cropped from the original images to accommodate the memory constraints of the computer hardware. Patches overlap by 256 pixels and their segmentation maps are stitched using a consensus algorithm[5].

Instance segmentation represents the most computationally demanding aspect of inference within AMAP as it requires clustering of pixel embeddings into separate foot process instances. AMAP executes the clustering algorithm for various numbers of clusters and relies on the silhouette score[10] to select the optimal cluster. Instead of clustering, AMAP-APP implements a postprocessing pipeline to derive the instance segmentation mask from the semantic segmentation output. After stitching all patches and constructing the semantic mask, AMAP-APP relabels slit diaphragm pixels as background. Next, the resulting binary mask is converted to an instance segmentation mask using a connected component labeling algorithm[11,12]. This procedure achieves outcomes similar to AMAP's instance segmentation while remaining computationally efficient.

**ROI Detection**

The enhanced ROI algorithm in AMAP-APP operates on the semantic segmentation output from the deep learning model.

The algorithm begins by up-sampling the segmentation map to the original image's full resolution using nearest-neighbor interpolation. A binary mask is then generated from this map, isolating the slit diaphragm regions while excluding foot process areas.

This initial SD mask is subjected to a series of morphological operations to define the ROI. First, a substantial morphological dilation is performed over many iterations using a large, circular structuring element. This operation massively expands and merges the distinct SD structures into large, continuous components, effectively encompassing the entire podocyte network. Following this expansion, a slight morphological erosion is applied using a small, cross-shaped kernel to smooth the boundaries of the resulting components.

The final step involves refining this mask by analysis of its connected components. All distinct regions in the mask are identified, and their areas are calculated. A size-based filter is then applied, and all components with an area smaller than a predefined minimum threshold are discarded. This filtering step effectively removes noise and ensures that only the main, contiguous podocyte regions are retained as the final ROI.

## Results

We evaluated the performance of AMAP-APP by benchmarking its computational efficiency and validating its analytical accuracy against the original AMAP method using both mouse and human datasets. Our benchmarks demonstrate that AMAP-APP achieves a substantial, approximately 147-fold increase in processing speed compared to the original GPU-accelerated implementation, effectively enabling high-throughput analysis on consumer-grade hardware. Despite this optimized processing pipeline, the application maintains high correlation ($r > 0.90$) and statistical equivalence to the original AMAP method for all quantified morphometric features, including foot process area, perimeter, and circularity. Furthermore, the implementation of a new Region of Interest (ROI) algorithm exhibited reduced deviation from manual delineations compared to the previous algorithm, resulting in improved precision for slit diaphragm length density measurements.

**Equivalency of AMAP-APP to AMAP on mouse data**

To assess the performance of AMAP-APP, we conducted a comparative analysis against the original AMAP method, utilizing a mouse kidney dataset. This dataset comprises 175 STED images, spanning various mouse ages and degrees of podocyte effacement. For this dataset, a semi-manual method, referred to as "Macro" served as an additional method for comparison. AMAP was compared to the Macro method in the original paper[5].

We evaluated the agreement and correlation of AMAP and AMAP-APP with the Macro method for key morphometric features: FP Area, Perimeter, and Circularity.

Table 1 presents the Pearson correlation coefficients for all features across different method pairing. AMAP-APP demonstrated a higher correlation with the Macro for area, slightly higher correlation for Perimeter and slightly lower for Circularity. Furthermore, a high correlation was observed between AMAP and AMAP-APP for all features, confirming the computational re-implementation preserves the core analytical outcomes.

Moreover, to formally test the practical equivalence of these methods, Two One-Sided T-tests (TOST)[13] were performed for all morphometric features, with an equivalence margin set at 10%. The TOST results demonstrated the statistical equivalence of AMAP and AMAP-APP for all evaluated features (Table 2), supporting that AMAP-APP provides comparable quantitative results to the original AMAP within an acceptable margin.

Figure 2 visualizes the agreement between AMAP and AMAP-APP in the Bland-Altman plots[14]. These plots illustrate the mean difference and limits of agreement, providing insights into systematic biases and variability.

**The new region of interest algorithm reduces deviation from manual delineations and enhances SD length measurements precision.**

Regarding the Region of Interest (ROI) algorithm, we specifically investigated their impact on slit diaphragm length density measurements. Both AMAP's original ROI algorithm and AMAP-APP's new ROI algorithm showed a high correlation with manually defined ROIs for SD length density; Previous ROI vs. Manual: $r = 0.9747$, $p < 0.0005$; New ROI vs. Manual: $r = 0.9644$, $p < 0.0005$. A Bland-Altman plot comparing the new ROI algorithm against manual ROI (Figure 3) further illustrates that the new algorithm exhibited reduced deviation from manual delineations, addressing the tendency of the original AMAP ROI algorithm to produce larger regions and thus enhancing the precision of SD length density measurements.

**AMAP-APP is applicable to clinical human patient material and conserves AMAP original performances.**

To demonstrate the broad applicability and consistency of AMAP-APP, we extended our comparative analysis to human kidney samples. The dataset comprises 190 STED images of human tissue, mirroring the human samples utilized in the original AMAP research. The comparison in this section focuses exclusively on the agreement and equivalence between AMAP and AMAP-APP.

We evaluated the consistency of AMAP-APP's measurements against those from AMAP for FP Area, Perimeter, Circularity.

Table 3 presents the Pearson correlation coefficients for all features comparing AMAP against AMAP-APP. Consistently high correlations were observed across all features, reinforcing the strong internal consistency and interchangeability of AMAP-APP with the original AMAP method for human samples.

To formally establish the practical equivalence of AMAP-APP's measurements to AMAP, TOST were conducted for all morphometric features, employing an equivalence margin of 10%. The results of the TOST analysis (Table 4) confirmed the statistical equivalence of AMAP-APP and AMAP for all evaluated features, indicating that within an acceptable margin AMAP-APP provides reliable and comparable morphometric quantifications for human podocytes.

Bland-Altman plots were also generated for each of these features (Figure 4, A-C) to visually assess the mean differences and the 95% limits of agreement, thereby illustrating the high level of agreement between the two automated methods.

**AMAP-APP is 147 faster than AMAP while conserving original performances.**

A critical objective of AMAP-APP was to significantly reduce the computational burden associated with the original AMAP method while maintaining analytical accuracy. To quantify this improvement, a benchmarking study was conducted, comparing the execution times of GPU-accelerated AMAP (amap-gpu) against AMAP-APP running on both GPU (amap-app-gpu) and CPU (amap-app-cpu). The tests were performed on a system equipped with an AMD Ryzen 9 7950X processor, an NVIDIA GeForce RTX 4090 with 24 GB of GDDR6X memory, and 128 GB of DDR5 RAM. The benchmarks were executed on sample images provided by AMAP's repository.

The results, depicted in Figure 5 (Mean Execution Times), demonstrate a substantial acceleration achieved by AMAP-APP. The original AMAP method, even when utilizing GPU acceleration (amap-gpu), required a mean execution time of 3213.95 seconds. In stark contrast, AMAP-APP exhibited dramatically reduced processing times. When running on the GPU (amap-app-gpu), AMAP-APP completed the task with a mean execution time of 21.82 seconds, representing an astonishing ~147-fold increase in speed compared to amap-gpu. Even when constrained to the CPU (amap-app-cpu), AMAP-APP maintained remarkable efficiency, with a mean execution time of 85.08 seconds, significantly outperforming the original GPU-accelerated AMAP.

AMAP can robustly reproduce quantification of food process effacement in the mouse model of focal segmental glomerulosclerosis and also is applicable to clinical human patient material.[3] Therefore the high correlation and statistical equivalence between the output of AMAP-APP and AMAP suggests that AMAP-APP can replace AMAP for quantitative analysis while being approximately 147 times faster.

## Discussion

Utilizing deep learning-based methods for the segmentation of foot processes and the slit diaphragm from super-resolution light microscopy images of podocytes enables non-biased and automated quantification of FP morphology at high throughput. AMAP previously introduced such a method, validated on both mouse and human sample images acquired using STED and confocal microscopy protocols. However, several limitations inherent to the original AMAP implementation — including the absence of a user interface, restricted operating system support (Linux-only), and the computationally intensive nature of its instance segmentation algorithm — significantly hindered the widespread availability and adoption of the method for scientific studies and potential clinical diagnostics. The original method's reliance on high-performance computing systems made it inaccessible to many researchers and clinical laboratories.

Here, we introduced AMAP-APP, an improved implementation designed as a cross-platform desktop application suitable for consumer-grade hardware. AMAP-APP expands usability by supporting Linux, macOS, and Windows. The core contribution of AMAP-APP is the replacement of AMAP's computationally expensive instance segmentation pipeline with a highly efficient classic image processing algorithm that operates directly on the semantic segmentation output. This postprocessing change, while retaining the same underlying neural network for semantic segmentation, drastically reduces computational requirements. As a result, AMAP-APP is approximately 147 times faster than AMAP, processing images in mere seconds compared to minutes, and critically, it can execute efficiently on both GPUs and standard consumer CPUs. This eliminates the need for specialized high-performance computing hardware, making the method accessible to a significantly broader user base.

We demonstrated through benchmarking and statistical equivalence testing that AMAP-APP provides morphometric quantifications for foot process's

Area, Perimeter, Circularity, and SD length density that are highly correlated with, and statistically equivalent to, those produced by the original AMAP method on both mouse and human datasets. A notable advancement is the introduction of a new ROI algorithm in AMAP-APP, which generates ROIs with less deviation from manual delineations, thereby improving the accuracy of slit diaphragm's length density measurements. Coupled with a simple and intuitive user interface offering project management capabilities, AMAP-APP not only overcomes previous hardware hurdles but also significantly improves ease of use.

In summary, AMAP-APP dramatically improves the availability and efficiency of deep learning-based methods for podocyte foot processes and slit diaphragm segmentation. By removing the stringent hardware requirements, offering broad platform compatibility, incorporating a more precise ROI algorithm, and providing a user-friendly interface, AMAP-APP democratizes this analytical method. It is open source and freely available to the broad community of podocyte research, fostering its adoption in both fundamental scientific investigations and ultimately contributing to advancements in clinical nephrology and diagnostics.

## Disclosure Statement

All the authors declared no competing interests.

# Acknowledgements

This work was supported by the Clinical Research Unit (CRU) 329 (KFO 329; A1, A6, and A7) as well as partly by Forschungsgruppe (FOR) 2743 of the Deutsche Forschungsgemeinschaft (DFG). KB was supported by the North Rhine-Westphalia return program (311-8.03.03.02-147635) and the Bundesministerium für Bildung und Forschung (BMBF) program Junior Group Consortia in Systems Medicine (01ZX1917B) and hosted by the Center for Molecular Medicine Cologne. LB was supported by the Koeln Fortune Program / Faculty of Medicine, University of Cologne. This work was partially funded by the Else Kröner-Fresenius-Stiftung and the Eva Luise und Horst Köhler Stiftung, project number 2019_KollegSE.04, the consortium STOP-FSGS by the German Ministry for Science and Education (BMBF01GM1901E to TB), Njurfonden (The Swedish Kidney Foundation, project numbers F2021-0004 and 2022-0051), and Torsten Söderbergs Stiftelse (project
number M64/21).

# Data Sharing Statement
The image data used in this study are openly available in Zenodo repository https://doi.org/10.5281/zenodo.13285467.
Additional data for method evaluation were obtained from the authors of the previous publication
https://www.nature.com/articles/s42255-020-0204-y.
The code used in the analyses is available in github at
https://github.com/bozeklab/amap-app.

## Table 1. Correlation Coefficients for Podocyte Morphometric Features Between AMAP and AMAP-APP in Mouse Kidney Tissue

| Feature | Comparison | Pearson's r |
|---|---|---|
| FP Area | AMAP vs. Macro | 0.6830 |
| | AMAP-APP vs. Macro | 0.7732 |
| | AMAP vs. AMAP-APP | 0.9080 |
| FP Perimeter | AMAP vs. Macro | 0.8490 |
| | AMAP-APP vs. Macro | 0.8556 |
| | AMAP vs. AMAP-APP | 0.9243 |
| FP Circularity | AMAP vs. Macro | 0.9638 |
| | AMAP-APP vs. Macro | 0.9469 |
| | AMAP vs. AMAP-APP | 0.9602 |

Pearson Correlation Coefficients across different method pairings for Podocyte Morphometric Features (Mouse Data). For all the reported coefficients: p-value was less than 0.0005.

## Table 2. Equivalence Testing of Mouse Podocyte Morphometric Features: AMAP-APP vs. AMAP

| Metric | Reference Mean (AMAP-APP) | Equivalence Margin (Absolute) | Mean Difference (AMAP-APP - AMAP) | Equivalence Declared |
|---|---|---|---|---|
| FP Area | 0.18 | +/- 0.02 | 0.00 | True |
| FP Perimeter | 2.10 | +/- 0.21 | 0.02 | True |
| FP Circularity | 0.50 | +/- 0.05 | 0.00 | True |

Table 2 summarizes the equivalence testing results for key podocyte morphometric features comparing AMAP-APP to AMAP, using a 10% equivalence margin ($\alpha=0.05$). Equivalence is declared if the TOST p-value is less than 0.05. For all reported rows p-value was less than 0.0005.

## Table 3. Correlation Coefficients for Podocyte Morphometric Features Between AMAP and AMAP-APP in Human Kidney Tissue

| Feature | Comparison | Pearson's r |
|---|---|---|
| FP Area | AMAP vs. AMAP-APP | 0.9244 |
| FP Perimeter | AMAP vs. AMAP-APP | 0.9424 |
| FP Circularity | AMAP vs. AMAP-APP | 0.8056 |

Table 3 demonstrates the strong correlation between the original AMAP method and AMAP-APP when applied to human kidney tissue. For FP Area and FP Perimeter, very high correlation coefficients (r > 0.92) were observed, indicating excellent agreement. FP Circularity also showed a strong correlation (r = 0.8056). For all the reported coefficients: p-value was less than 0.0005.

## Table 4. TOST Analysis for Human Podocyte Morphometric Equivalence: AMAP-APP vs. AMAP

| Feature | Equivalence Bounds (10% of AMAP mean) | Mean Difference (AMAP-APP - AMAP) | 95% Confidence Interval (CI) | Conclusion |
|---|---|---|---|---|
| FP Area | [-0.0108, 0.0108] | 0.0092 | [0.0078, 0.0107] | Equivalent |
| FP Perimeter | [-0.1392, 0.1392] | 0.0950 | [0.0860, 0.1040] | Equivalent |
| FP Circularity | [-0.0592, 0.0592] | -0.0066 | [-0.0093, -0.0039] | Equivalent |

Table 4 details the results of the TOST analysis, which assessed the statistical equivalence of AMAP-APP measurements compared to those from the original AMAP for human podocyte morphometry. For all evaluated features—FP Area, FP Perimeter, and FP Circularity—the TOST p-values were consistently below the significance level of 0.05 (with an equivalence margin set at 10% of the AMAP mean). TOST p-value for FP Area was 0.0167 and for the rest it was less than 0.0005.

# Figure Legend

## Figure 1. AMAP-APP Architecture

This figure illustrates the modified U-Net architecture[5]. The network generates a 3-channel semantic segmentation mask and a 16-channel embedding output. While the original AMAP utilized both outputs, AMAP-APP discards the computationally expensive embeddings and instead employs a classic image processing pipeline on the semantic mask for efficient instance segmentation.

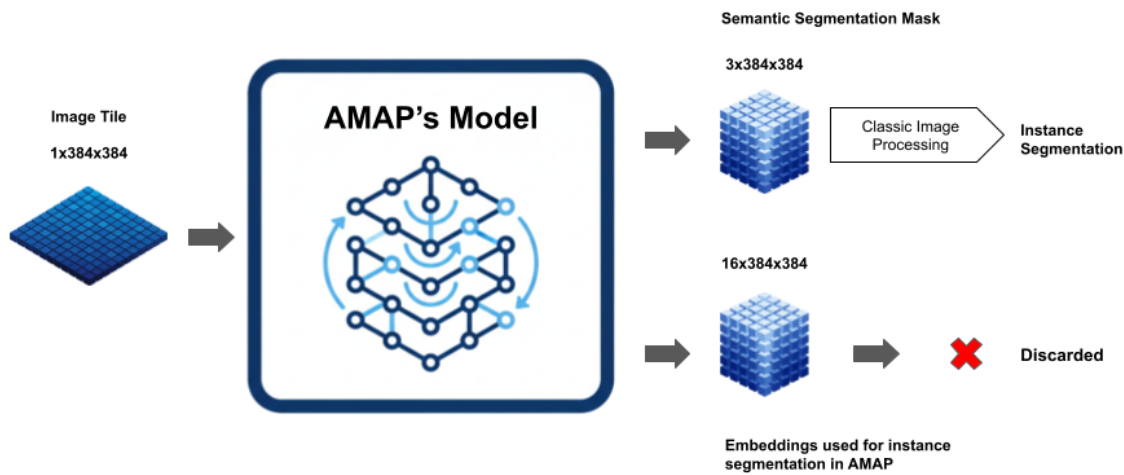

## Figure 2. Bland-Altman Plots Comparing AMAP and AMAP-APP for Key Podocyte Morphometric Features.

This figure presents Bland-Altman plots assessing the agreement between the original AMAP method and AMAP-APP for foot process (FP) morphometry. **(A)** FP Area: The mean difference between AMAP and AMAP-APP is -0.00, with 95% limits of agreement (LoA) from -0.04 to 0.04. **(B)** FP Circularity: The mean difference is -0.00, with 95% LoA from -0.05 to 0.05. **(C)** FP Perimeter: The mean difference is -0.02, with 95% LoA from -0.36 to 0.31. Across all three features, the mean differences are near zero and the majority of data points fall within the 95% LoA, indicating strong agreement and interchangeability between AMAP and AMAP-APP.

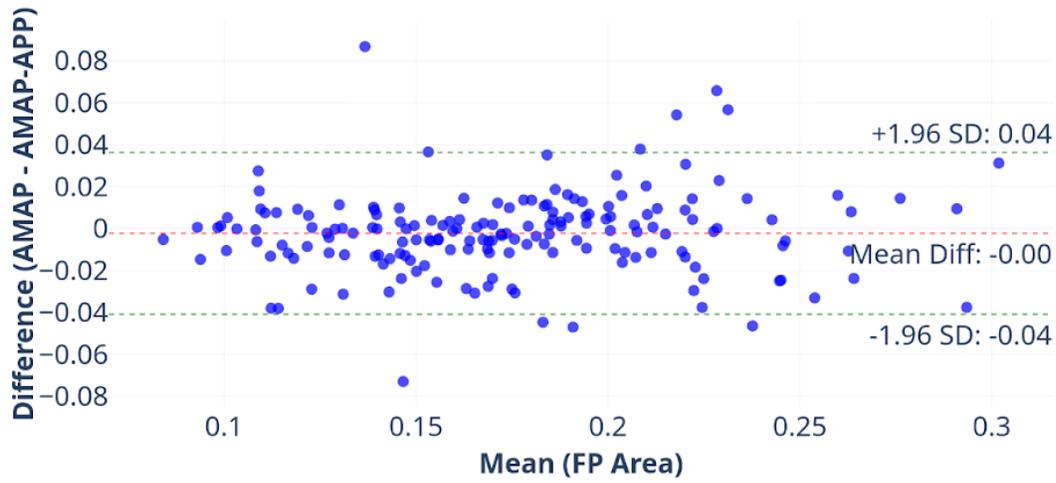

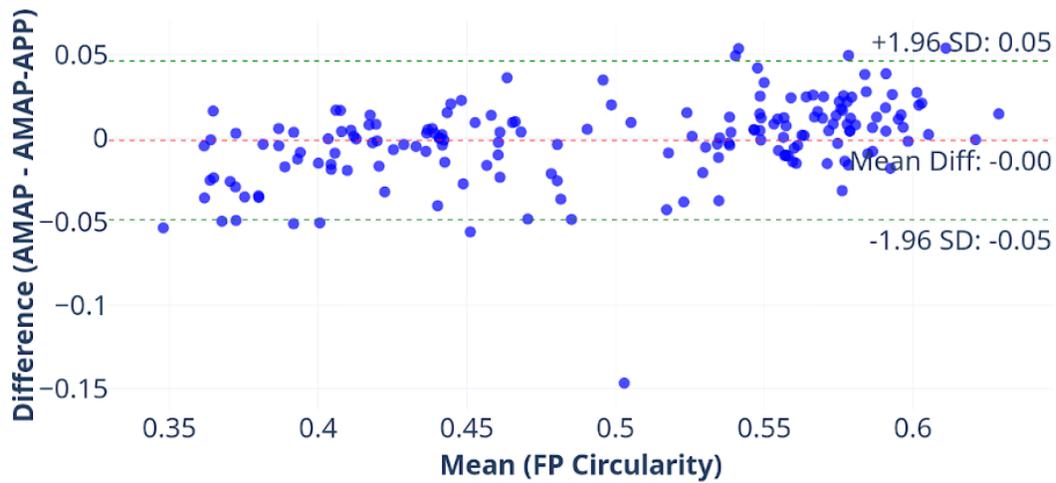

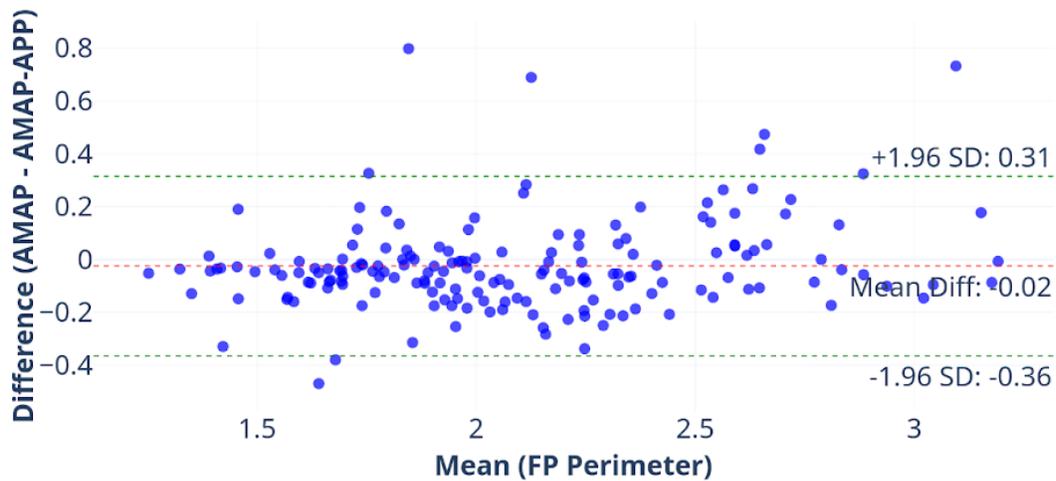

## Figure 3. Bland-Altman Plots Comparing SD Length Density from ROI Algorithms against Macro.

**(A)** Bland-Altman plot illustrating the agreement between the Macro and AMAP-APP's new ROI algorithm for Slit Diaphragm (SD) length density. The mean difference is 0.20, with 95% limits of agreement ranging from -0.65 to 1.06. **(B)** Bland-Altman plot comparing the Macro method with AMAP's original (old) ROI algorithm for SD length density. The mean difference is 0.59, with 95% limits of agreement ranging from -0.13 to 1.31. The narrower limits of agreement and smaller mean difference in (A) indicate superior agreement and less deviation of the new ROI algorithm compared to the old one against the Macro method.

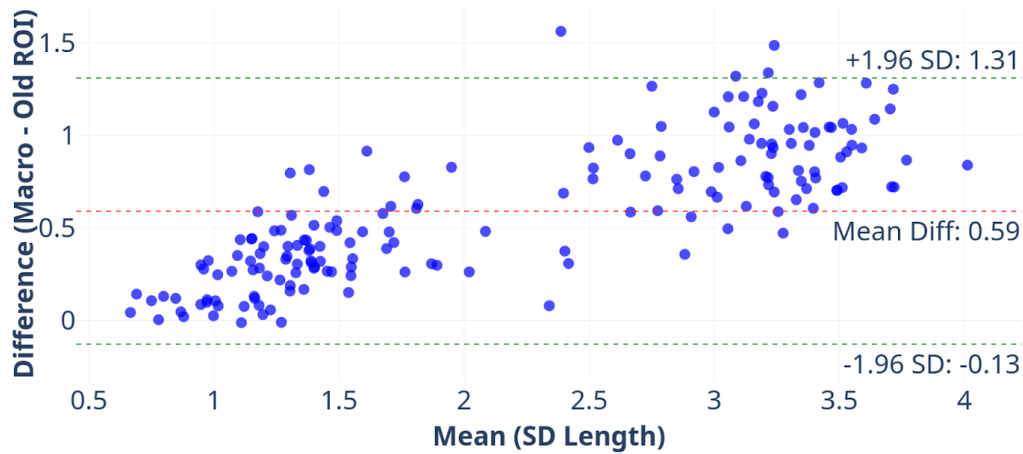

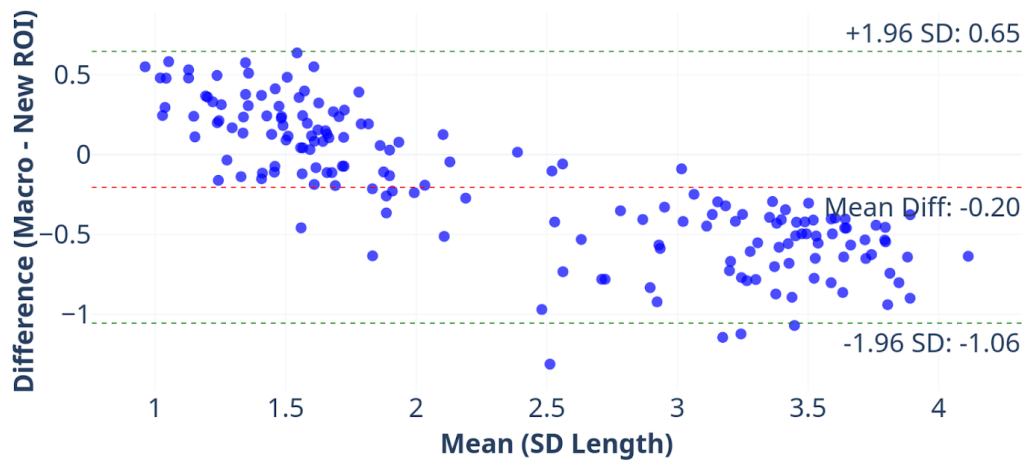

# Figure 4. Bland-Altman Plots for Podocyte Morphometric Features: AMAP vs. AMAP-APP (Human Data).

This figure presents Bland-Altman plots assessing the agreement between the original AMAP method and AMAP-APP for foot process (FP) morphometry on the human dataset. **(A) FP Area:** The plot shows a bias of 0.01, with 95% limits of agreement (LoA) ranging from -0.01 to 0.03. **(B) FP Circularity:** The bias is -0.01, with 95% LoA ranging from -0.04 to 0.03. **(C) FP Perimeter:** The bias is 0.09, with 95% LoA ranging from -0.03 to 0.22. These plots collectively demonstrate a high level of agreement between AMAP and AMAP-APP, and the majority of data points falling within the narrow 95% LoA for all evaluated features in human samples.

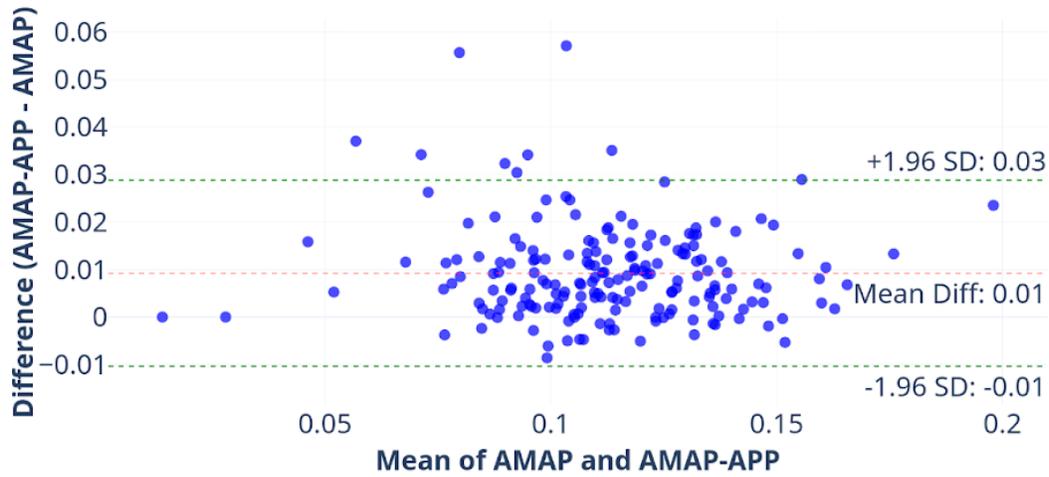

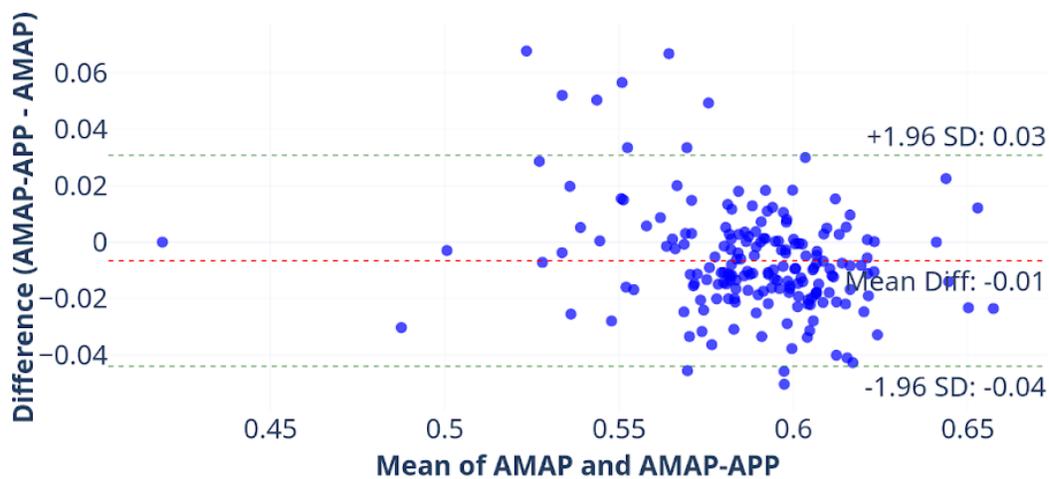

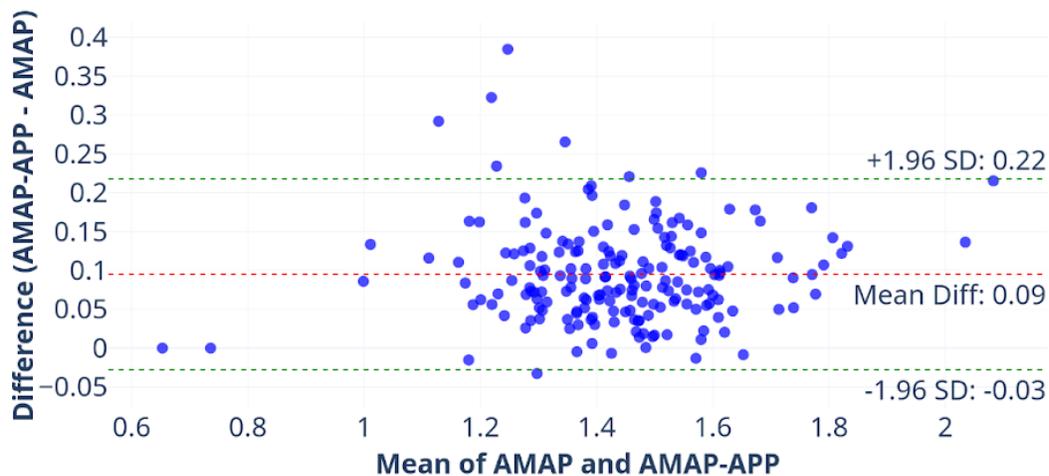

# Figure 5. Execution Times per Iteration
further illustrates the consistent performance gains across individual processing iterations.

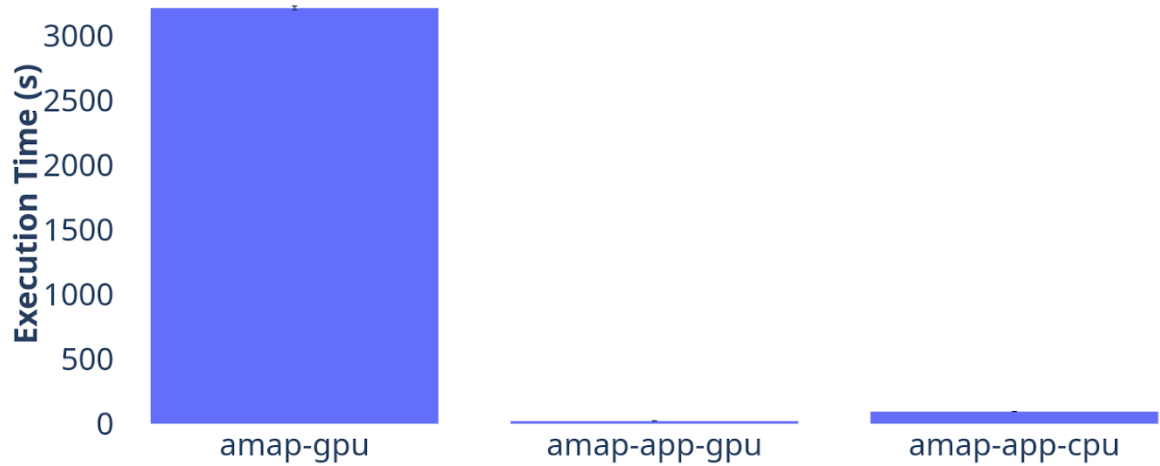

a) Mean Execution Times

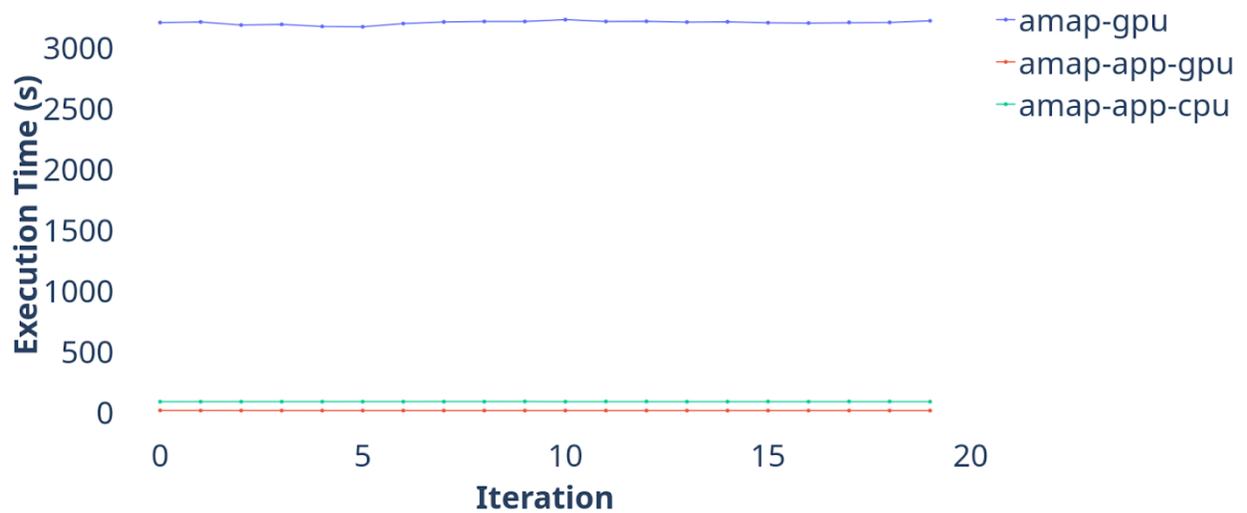

b) Execution Times per Iteration

# Figure 6. Results of AMAP-APP with new and old ROI algorithm

This figure compares the performance of the initial ROI selection algorithm (top row) against the updated algorithm incorporated into the current AMAP-APP (bottom row) on three tissue samples. The older method generated broader ROIs that frequently included. The improved algorithm achieves much tighter segmentation, closely contouring the actual boundaries.

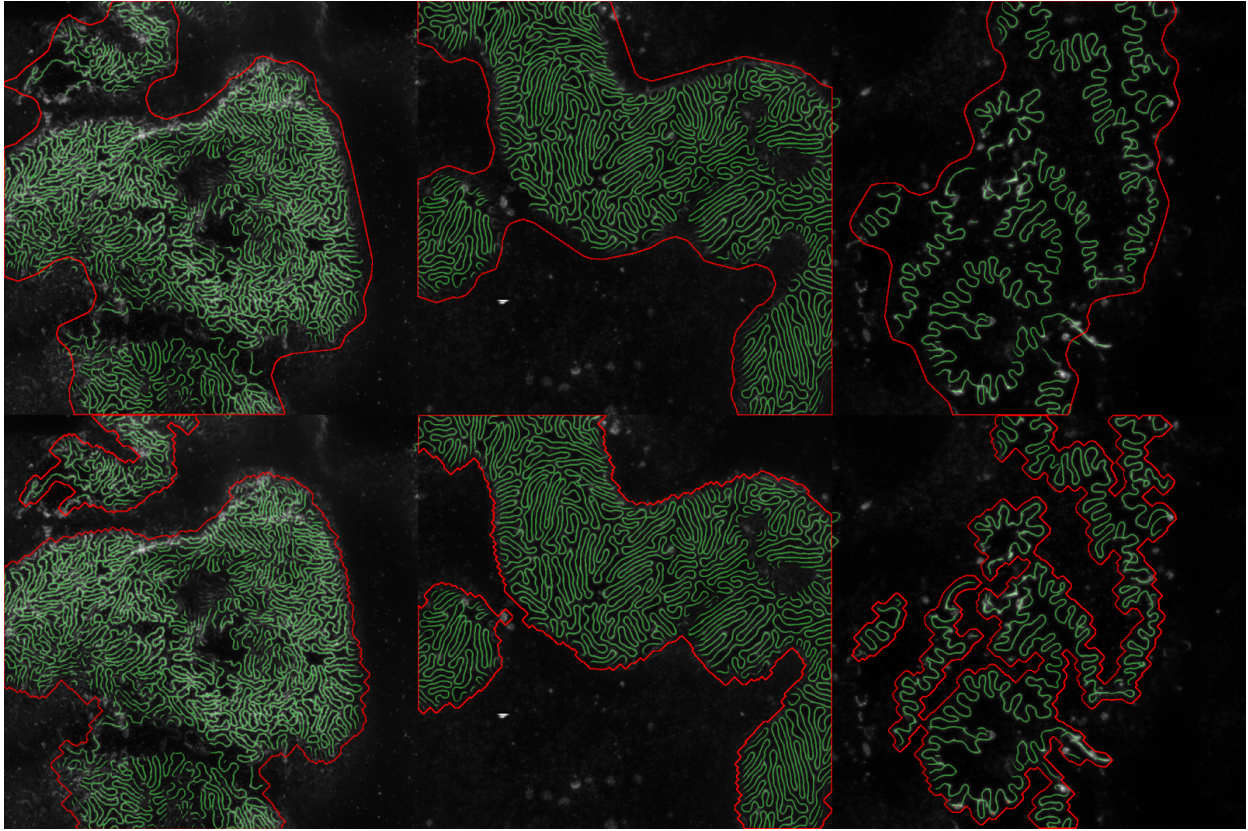

# Figure 7. Representative Visualization of AMAP-APP Segmentation on Human Tissue Samples.

This panel displays three sets of human podocyte samples processed using AMAP-APP to demonstrate its applicability to clinical material. Left Column: Instance Segmentation Output, where individual foot process instances are identified and labeled with distinct colors to allow for individual morphometric quantification. Middle Column: Semantic Segmentation Output, showing the categorical classification of pixels as foot processes against the background. Right Column: Slit Diaphragm Detection, highlighting the identified slit diaphragm structures within the automated ROI (red boundary) used for length density calculations.

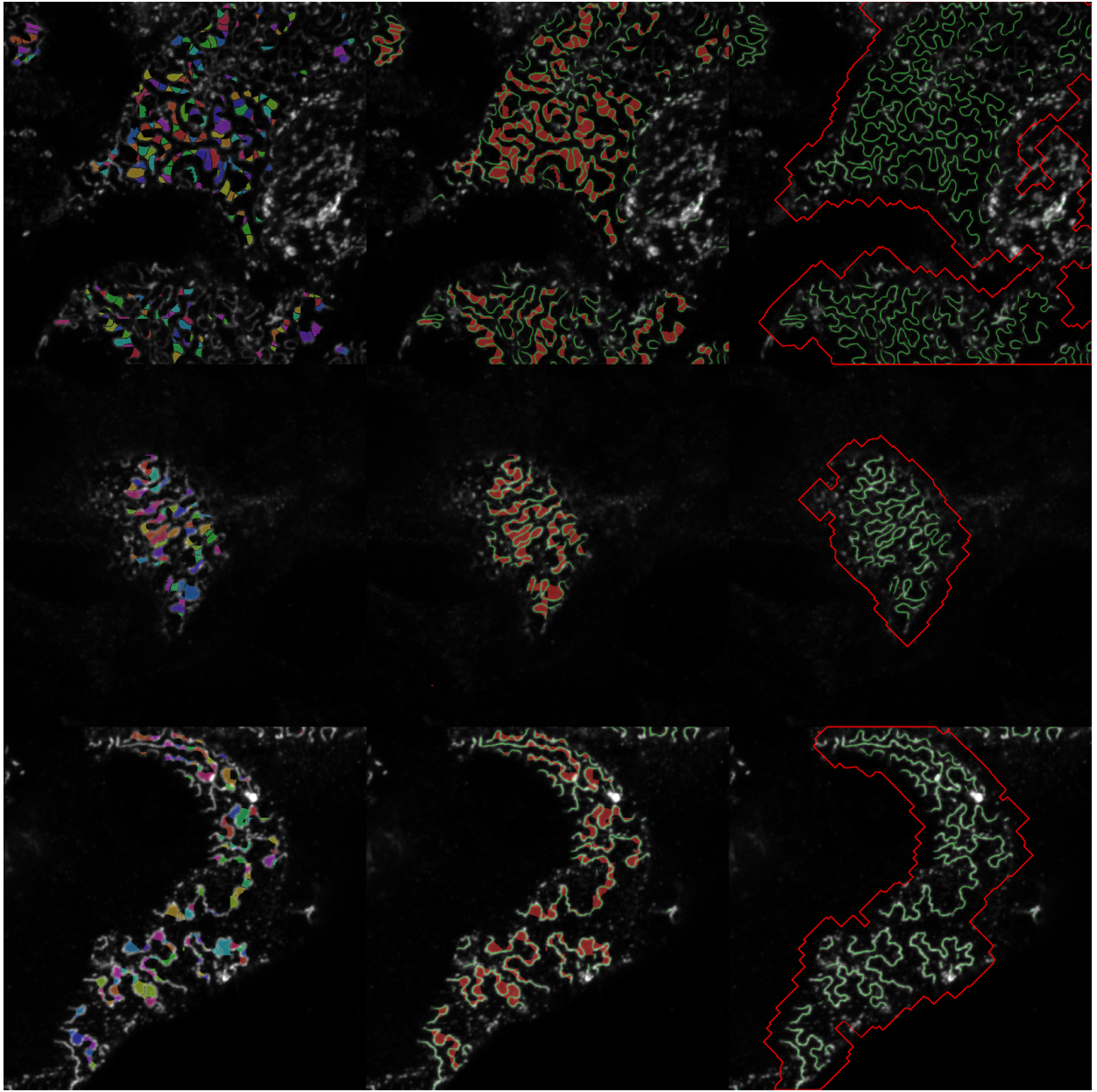